# Representation Policy Iteration

**Sridhar Mahadevan**
Department of Computer Science
University of Massachusetts
140 Governor's Drive
Amherst, MA 01003
mahadeva@cs.umass.edu

## Abstract

This paper addresses a fundamental issue central to approximation methods for solving large Markov decision processes (MDPs): how to automatically learn the underlying representation for value function approximation? A novel theoretically rigorous framework is proposed that automatically generates *geometrically customized* orthonormal sets of basis functions, which can be used with any approximate MDP solver like least-squares policy iteration (LSPI). The key innovation is a *coordinate-free* representation of value functions, using the theory of smooth functions on a Riemannian manifold. Hodge theory yields a constructive method for generating basis functions for approximating value functions based on the eigenfunctions of the self-adjoint (Laplace-Beltrami) operator on manifolds. In effect, this approach performs a global Fourier analysis on the state space graph to approximate value functions, where the basis functions reflect the large-scale topology of the underlying state space. A new class of algorithms called Representation Policy Iteration (RPI) are presented that automatically learn both basis functions and approximately optimal policies. Illustrative experiments compare the performance of RPI with that of LSPI using two handcoded basis functions (RBF and polynomial state encodings).

## 1 Introduction

This paper presents a unified framework for learning representation and behavior, integrating policy learning using the framework of *Markov decision processes* [16] with automated representation discovery using abstract harmonic analysis to build geometrically customized basis functions on graphs [4]. The resulting combined approach enables agents to learn representations that reflect both their past experience and the inherent large-scale geometry of the environment, as well as learn policies by using these customized basis functions to approximate value functions. The key innovation is a *coordinate-free* representation of value functions, using the Hilbert space of smooth functions on a Riemannian manifold. This approach allows applying powerful mathematical tools for basis function generation by exploiting the properties of the self-adjoint Laplace operator on Riemannian manifolds. The Laplace (or Laplace-Beltrami) operator $\Delta$ is a positive-semidefinite self-adjoint operator on differentiable functions. Hodge theory shows that the Hilbert space of smooth functions has a discrete spectrum on compact manifolds, captured by the eigenfunctions of the Laplacian [17]. The Laplacian operator acts on smooth functions on a manifold (e.g,. a graph) in a way analogous to the Bellman operator on value functions: both enforce geodesic smoothing.

Markov decision processes (MDPs) are a well-studied model of acting under uncertainty studied in decision-theoretic planning [3] and reinforcement learning [19]. A finite MDP $M = \langle S, A, P^a_{ss'}, R^a_{ss'} \rangle$ is specified by a set of states $S$, a set of actions $A$, a transition model $P^a_{ss'}$ specifying the distribution over future states $s'$ when an action $a$ is performed in state $s$, and a corresponding reward model $R^a_{ss'}$ specifying a scalar cost or reward. Methods for solving MDPs have been studied for almost 50 years: exact methods such as *value iteration* and *policy iteration* represent value functions using a table [16]; approximate methods like least-squares policy iteration (LSPI) and linear programming methods project the exact value function onto a subspace spanned by a set of handcoded basis functions [6, 7, 8]. The proposed approach can be viewed as automatically generating subspaces for projecting the value function using the framework of harmonic analysis [1]. The framework for representation discovery is based on spectral analysis of a self-adjoint symmetric random walk operator on graphs [4]. The proposed approach fundamentally differs from past work on basis function generation, for example tuning a pre-defined set of functions [13] or generating tabular basis functions for a linear programming based value function approximator using a greedy algorithm, based on the error in approximating a particular value function [15].

## 2 Overview of the Approach

The framework combines Samuel's value function paradigm with Amarel's representation learning paradigm [11]. The framework also provides a new formulation of task-independent RL, where agents learn global basis functions or *proto-value functions* [10]. In this paper, we present the framework as an enhancement of policy iteration, namely representation policy iteration (RPI), since it enables learning both policies and the underlying representations. The proposed framework uses spectral graph theory [4] to build basis representations for smooth (value) functions on graphs induced by Markov decision processes. Any policy in an MDP can be viewed as a Markov chain partitioning states into transient and recurrent states. The construction of basis functions is based on analyzing a symmetric operator (the graph Laplacian) which can be viewed as a reversible random walk on the state space. It is important to note that the set of basis functions learned from a reversible random walk is still useful in approximating value functions for any policy. Thus, the process of learning basis functions by spectral analysis of reversible random walks is an *off-policy* representation learning method, in the sense that the actual MDP dynamics under a specific policy may induce a non-reversible random walk. Reversible random walks greatly simplify spectral analysis since such random walks are similar to a symmetric operator on the state space.

For any graph $G$, the "tabular" representation is the orthonormal set of basis functions $\phi(i) = [0 \ldots i \ldots 0]$, one for each vertex (state) in the graph. Besides being inefficient, this representation does not exploit the topology of the specific graph. Intuitively, a representation should reflect the (irregular) topology of the state space under consideration. This problem also afflicts more efficient approximations, such as the polynomial basis functions where $\phi(s) = [1 \ s \ldots s^k]$ for some fixed $k$ [8]. While this encoding is sparse, it is numerically unstable for large graphs, dependent on the ordering of vertices, and once again, insensitive to the underlying state space geometry. Our approach addresses these shortcomings by building basis functions on graphs using harmonic analysis of the graph topology. First, we briefly review linear approximation methods for solving MDPs, in particular LSPI. LSPI and other linear approximation methods approximate the true action-value function $Q^\pi(s, a)$ for a policy $\pi$ using a set of handcoded basis functions $\phi(s, a)$ that can be viewed as doing dimensionality reduction: the true action value function $Q^\pi(s, a)$ is a vector in a high dimensional space $\mathcal{R}^{|S| \times |A|}$, and using the basis functions amounts to reducing the dimension to $\mathcal{R}^k$ where $k \ll |S| \times |A|$. The approximated action value is thus

$$\hat{Q}^\pi(s, a; w) = \sum_{j=1}^{k} \phi_j(s, a) w_j$$

where the $w_j$ are weights or parameters that can be determined using a least-squares method. Let $Q^\pi$ be a real (column) vector $\in \mathcal{R}^{|S| \times |A|}$. The column vector $\phi(s, a)$ is a real vector of size $k$ where each entry corresponds to the basis function $\phi_j(s, a)$ evaluated at the state action pair $(s, a)$. The approximate action-value function can be written as $\hat{Q}^\pi = \Phi w^\pi$, where $w^\pi$ is a real column vector of length $k$ and $\Phi$ is a real matrix with $|S| \times |A|$ rows and $k$ columns. Each row of $\Phi$ specifies all the basis functions for a particular state action pair $(s, a)$, and each column represents the value of a particular basis function over all state action pairs. LSPI uses a least-squares fixed-point approximation $T_\pi Q^\pi \approx Q^\pi$, where $T_\pi$ is the Bellman backup operator. This latter approach yields the following solution for the coefficients:

$$w^\pi = \left(\Phi^T \left(\Phi - \gamma P \Pi_\pi \Phi\right)\right)^{-1} \Phi^T R$$

where $R$ is the reward vector (over all state action pairs), $P$ is the transition matrix, and $\Pi_\pi$ is a stochastic matrix of size $|S| \times |S||A|$ where $\Pi_\pi(s, (s, a)) = \pi(a; s)$. LSPI uses the LSTDQ (least-squares TD Q-learning) method as a subroutine for learning the state-action value function $\hat{Q}^\pi$. The LSTDQ method solves the system of linear equations $Aw^\pi = b$ where

$$A = \Phi^T \Delta_\mu \left(\Phi - \gamma P \Pi_\pi\right)$$

Also, $\mu$ is a probability distribution over $S \times A$ that defines the projection of the true action value function onto the subspace spanned by the handcoded basis functions, and $b = \Phi^T \Delta_\mu R$. Since $A$ and $b$ are unknown when learning, they are approximated from samples using the update equations

$$\begin{aligned} \tilde{A}^{t+1} &= \tilde{A}^t + \phi(s_t, a_t) \left(\phi(s_t, a_t) - \gamma \phi(s'_t, \pi(s'_t))\right)^T \\ \tilde{b}^{t+1} &= \tilde{b}^t + \phi(s_t, a_t) r_t \end{aligned}$$

where $(s_t, a_t, r_t, s'_t)$ is the $t^{th}$ sample of experience from a trajectory generated by the agent (using some random or guided policy), and $\tilde{A}^0$ and $\tilde{b}^0$ are set to 0.

LSTDQ computes the $\tilde{A}$ matrix and $\tilde{b}$ column vector, and then returns the coefficients $\tilde{w}^\pi$ by solving the system $\tilde{A}w^\pi = \tilde{b}$. The overall LSPI method uses a policy iteration procedure, starting with a policy $\pi$ defined by an initial weight vector $w$, and then repeatedly invoking LSTDQ to find the updated weights $w'$, and terminating when the difference $\|w - w'\| \leq \epsilon$.

## 3 Representation Policy Iteration

With this brief overview of the framework and linear least-squares approximation methods such as LSPI, we can now introduce the Representation Policy Iteration framework, which interleaves representation learning and policy learning. For ease of exposition, we present the complete algorithm in Figure 1 in the context of LSPI and illustrate its behavior, before providing a detailed overview of the underlying theory. Also, much of this introductory paper will focus on explaining the theory of how representations can be learned by harmonic analysis of a given set of sample transitions. It is clear that if representations can be successfully learned from an experience sample, then representation learning can easily be interleaved with policy learning by generating a new sample of transitions from executing the current policy. A detailed algorithmic analysis of fully interleaved policy and representation learning is beyond the scope of this introductory paper.

---

Representation Policy Iteration$(D_0, \gamma, k, \epsilon, \pi_0)$:

// D: Source of samples (s, a, r, s')
// $\gamma$: Discount factor
// $\epsilon$: Stopping criterion
// $\pi_o$: Initial policy specified as a weight vector $w_0$.
// $k$: number of (unknown) basis functions

1. Use the initial source of samples $D_0$ to construct the basis functions $\phi_1^0, \ldots, \phi_k^0$ as follows:
   (a) Build an approximation of the underlying manifold $\mathcal{M}$, e.g. using an undirected or directed neighborhood graph $G$ that encodes the underlying state (action) space topology.
   (b) Compute the lowest-order $k$ eigenfunctions $\psi_1, \ldots, \psi_k$ of the (combinatorial or normalized) graph Laplacian operator on the graph $G$. Obtain the basis functions $\phi_i^0$ from the eigenfunctions $\psi_i$ by repeating the state encoding $|A|$ times, and multiplying each eigenfunction $\psi_i$ by the indicator function $I(a_i = a)$.

2. $\pi' \leftarrow \pi_0$.      // $w \leftarrow w_0$

3. **repeat**
   (a) $\pi_t \leftarrow \pi'$.      // $w \leftarrow w'$
   (b) **Optional**: compute a new set of basis functions $\phi^t$ by generating a new sample $D_t$ by executing $\pi_t$ and repeating step 1.
   (c) $\pi' \leftarrow \text{LSTDQ}(D, k, \phi^t, \gamma, \pi)$
   (d) $t \leftarrow t + 1$

4. **until** $\pi \sim \pi'$      // $\|w - w'\| \leq \epsilon$

Figure 1: Representation Policy Iteration interleaves representation and policy learning.

---

Steps 1 and 3b automatically build customized basis functions given an experience sample, where the basis functions reflect large-scale geometry of the approximated manifold, and provide a theoretically guaranteed orthonormal set for approximating any smooth function on the approximated manifold.

### 3.1 Abstract Harmonic Analysis

Temporarily setting aside the theory underlying RPI (see the sections below), we illustrate its performance on the classic chain example from [7, 8]. The chain MDP, originally studied in [7], is a sequential open chain of varying number of states, where there are two actions for moving left or right along the chain. The reward structure can vary, such as rewarding the agent for visiting the middle states, or the end states. Instead of using a fixed state action encoding, our approach automatically derives a customized encoding that reflects the *topology* of the chain. This can be viewed as doing an abstract harmonic analysis [1] of the graph encoding the topology. This process can be applied to any MDP: Figure 2 shows the basis functions that are created for an open and closed chain, and Figure 6 illustrates some of the basis functions from abstract harmonic analysis on a five-room grid world.

Abstract harmonic analysis [1] generalizes fixed orthonormal basis sets such as polynomial encodings. If we orthonormalize the monomials $\phi(s) = [1 \ s \ \ldots s^i]^T$ over $[-1, 1]$, we obtain a basis set of polynomials for $C_2[-1, 1]$, the set of all continuous functions over $[-1, 1]$, namely the well-known *Legendre* polynomials, whose first three items are $p_0(t) = \frac{1}{\sqrt{2}}$, $p_1(t) = \frac{\sqrt{6}}{2}t$, and $p_2(t) = \frac{\sqrt{10}}{4}(3t^2 - 1)$. If a weighted space is chosen, where $w(t) = \frac{1}{\sqrt{(1-t^2)}}$, this process yields the *Chebyshev polynomials*. Finally, if the underlying set is discrete, where $T = \{t_1, \ldots, t_m\}$ and $t_i = i$, then the resulting orthonormal set of polynomials spans the space $l_2(m)$. In RPI, the orthonormal set of eigenfunctions is built from modeling the underlying manifold. The eigenfunctions of the Laplace-Beltrami operator on a graph not only reflect the large-scale difference in the *topology* of the open or closed chain, they also can approximate any real-valued smooth function on the vertices of this graph. In the limit, Hodge theory (see below) guarantees that the approximation will be exact.

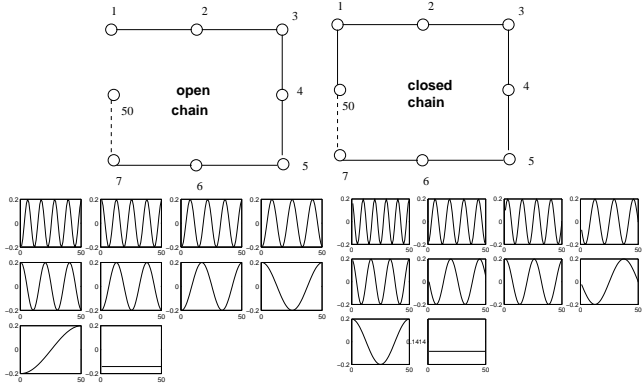

Figure 2: Laplacian eigenfunctions for a 50 state open and closed chain MDP produced from a random walk of 10000 steps by learning the underlying graph and computing its combinatorial graph Laplacian.

Figure 3 and Figure 4 shows the results of running the Representation Policy Iteration (RPI) algorithm on a 50 node chain graph, using the display format from [8]. Here, being in states 10 and 41 earns the agent rewards of +1 and there is no reward otherwise. The optimal policy is to go right in states 1 through 9 and 26 through 41 and left in states 11 through 25 and 42 through 50. The number of samples initially collected was set at $10,000$. The discount factor was set at $\gamma = 0.8$. By increasing the number of desired basis functions, it is possible to get very accurate approximation, although as Figure 4 shows, even a crude approximation using 5 basis functions is sufficient to learn a close to optimal policy. Using 20 basis functions, the learned policy is exact. A detailed comparison with handcoded polynomial and RBF encoding techniques studied earlier with LSPI is given in Section 6.

## 4 Spectral Graph Theory

We provide a brief theoretical overview of spectral graph theory and Riemannian manifolds. While these subfields have been applied to semi-supervised learning [2], clustering [14], and image segmentation [18], their use for representation learning and value function approximation in MDPs and RL is novel, which unlike supervised learning or classification requires approximating arbitrary real-valued functions. The graph Laplacian [4] can be used to construct a set of orthonormal basis functions $\phi_1^G(s), \ldots, \phi_k^G(s)$ that in the limit are guaranteed to exactly capture any smooth real-valued function on $G$. The graph Laplacian is a special case of the Laplacian on a Riemannian manifold, and this result generalizes to all connected oriented Riemannian manifolds [17]. But, let us first focus on the simple case of (connected) undirected graphs. Let $d_v$ denote the degree of vertex $v$. The random walk operator on a graph is given by $D^{-1}A$, where $D$ is the diagonal matrix whose entries are given by the vertex degrees, or row sums of the adjacency matrix $A$. The random walk operator is not symmetric, but it is closely related to a symmetric graph Laplacian operator as follows. The *combinatorial Laplacian* $L$ is defined as $D - A$, and acts on a function $f$ as

$$Lf(x) = \sum_{y \sim x}(f(x) - f(y))$$

where $y \sim x$ means $y$ is adjacent to $x$. It is easy to show that the combinatorial Laplacian is essentially the discrete case of the well-known Laplacian partial differential equation on a graph. The *nor-* malized Laplacian $\mathcal{L}$ of the graph $G$ is defined as $D^{-\frac{1}{2}}(D-A)D^{-\frac{1}{2}}$ or in more detail

$$\mathcal{L}(u,v) = \begin{cases} 1 & \text{if } u = v \text{ and } d_v \neq 0 \\ -\frac{1}{\sqrt{d_u d_v}} & \text{if } u \text{ and } v \text{ are adjacent} \\ 0 & \text{otherwise} \end{cases}$$

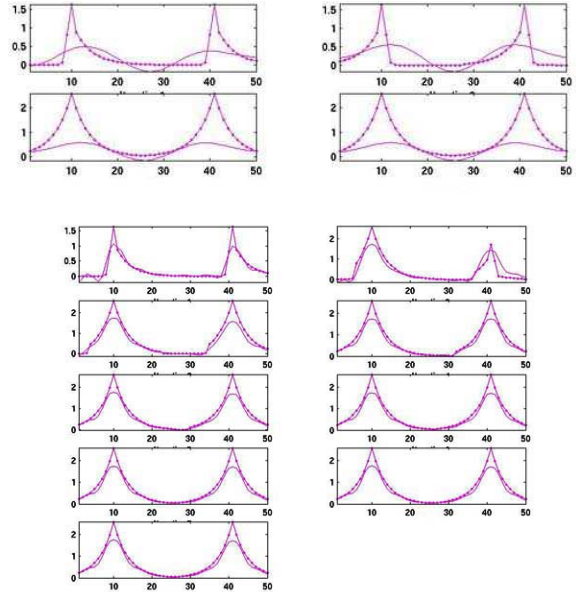

Figure 3: Representation Policy Iteration on a 50 node chain graph, for $k = 5$ basis functions (top four panels) and $k = 20$ (bottom nine panels). Each group of plots shows the state value function for each iteration (in row major order) over the 50 states, where the solid curve is the approximation and the dotted lines specify the exact function. Notice how the value function approximation gets much better at $k = 20$, as Hodge theory guarantees asymptotic convergence. Although the approximation is relatively poor at $k = 5$, the policy learned turns out to be close to optimal.

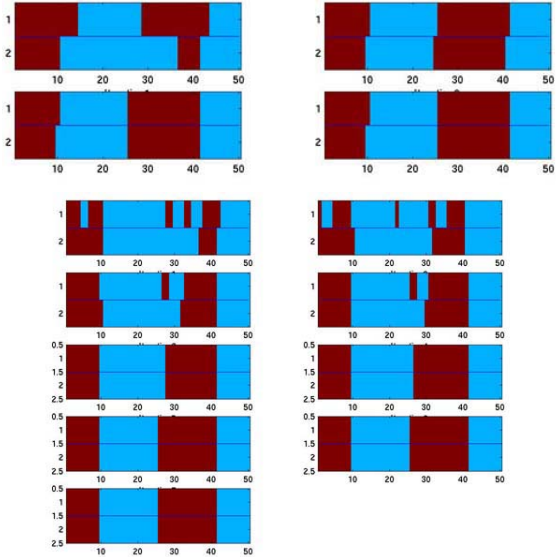

Figure 4: The policies learned at each iteration using Representation Policy Iteration on a 50 node chain graph, for 5 basis functions (top four panels), and 20 basis functions (bottom nine panels) in row major order. Even using 5 basis functions results in a close to optimal policy. The light (blue) color denotes the left action and the dark (red) denotes going right. The top half of each plot is exact, and the bottom is approximate.

It can be shown that $\mathcal{L}$ is a symmetric self-adjoint operator, and its spectrum (eigenvalues) lie in the interval $(0, 2)$. If $G$ is a constant degree $k$ graph, then it follows that $\mathcal{L} = I - \frac{1}{k}A$, where $A$ is the adjacency matrix of $G$. For a general graph $G$, $\mathcal{L} = D^{-\frac{1}{2}}LD^{-\frac{1}{2}} = I - D^{-\frac{1}{2}}AD^{-\frac{1}{2}}$. Note that this implies that $D^{-1}A = D^{-\frac{1}{2}}(I - \mathcal{L})D^{\frac{1}{2}}$. In other words, the random walk operator $D^{-1}A$ is similar to $I - \mathcal{L}$ in that both have the same eigenvalues, but the eigenfunctions of the random walk operator are the eigenfunctions of $I - \mathcal{L}$ scaled by $D^{-\frac{1}{2}}$. The Laplacian $\mathcal{L}$ is an operator on the space of functions defined on the graph $g: V \to \mathcal{R}$, where $V$ is the vertex set of $G$, and $u \sim v$ means $u$ and $v$ are neighbors:

$$\mathcal{L}g(u) = \frac{1}{\sqrt{d_u}} \sum_{v: u \sim v} \left( \frac{g(u)}{\sqrt{d_u}} - \frac{g(v)}{\sqrt{d_v}} \right) \quad (1)$$

The *Rayleigh* quotient provides a variational characterization of eigenvalues using projections of an arbitrary function $g: V \to \mathcal{R}$ onto the subspace $\mathcal{L}g$. The quotient gives the eigenvalues and the functions satisfying orthonormality are the eigenfunctions (here $\langle f, g \rangle = \sum_u f(u)g(u)$ denotes the inner product on graph $G$):

$$\frac{\langle g, \mathcal{L}g \rangle}{\langle g, g \rangle} = \frac{\langle g, D^{-\frac{1}{2}}LD^{-\frac{1}{2}}g \rangle}{\langle g, g \rangle} = \frac{\sum_{u \sim v}(f(u) - f(v))^2}{\sum_u f^2(u)d_u}$$

where $f \equiv D^{-\frac{1}{2}}g$. The first eigenvalue is $\lambda_0 = 0$, and is associated with the constant function $f(u) = \mathbf{1}$, which means the first eigenfunction $g_o(u) = \sqrt{D}\,\mathbf{1}$. The first eigenfunction (associated with eigenvalue 0) of the combinatorial Laplacian is the constant function $\mathbf{1}$. The second eigenfunction is the infimum over all functions $g: V \to \mathcal{R}$ that are perpendicular to $g_o(u)$, which gives us a formula to compute the first non-zero eigenvalue $\lambda_1$, namely

$$\lambda_1 = \inf_{f \perp \sqrt{T}\mathbf{1}} \frac{\sum_{u \sim v}(f(u) - f(v))^2}{\sum_u f^2(u)d_u}$$

The Rayleigh quotient for higher-order basis functions is similar: each function is perpendicular to the subspace spanned by previous functions. A further property of Laplacian eigenfunctions is that they transparently reflect the nonlinear geometry of state spaces (see experiments below). To formally explain this, we briefly review spectral geometry. The *Cheeger* constant $h_G$ of a graph $G$ is defined as

$$h_G(S) = \min_S \frac{|E(S, \tilde{S})|}{\min(\text{vol } S, \text{vol } \tilde{S})}$$

Here, $S$ is a subset of vertices, $\tilde{S}$ is the complement of $S$, and $E(S, \tilde{S})$ denotes the set of all edges $(u, v)$ such that $u \in S$ and $v \in \tilde{S}$. The volume of a subset $S$ is defined as $\text{vol } S = \sum_{x \in S} d_X$. Consider the problem of finding a subset $S$ of states such that the edge boundary $\partial S$ contains as few edges as possible, where $\partial S = \{(u,v) \in E(G) : u \in S \text{ and } v \notin S\}$. The relation between $\partial S$ and the Cheeger constant is given by $|\partial S| \geq h_G \text{ vol } S$ In the two-room grid world task illustrated below in Figure 5, the Cheeger constant is minimized by setting $S$ to be the states in the first room, since this will minimize the numerator $E(S, \tilde{S})$ and maximize the denominator $min(\text{vol } S, \text{vol } \tilde{S})$. A remarkable identity connects the Cheeger constant with the spectrum of the Laplace-Beltrami operator. This theorem underlies the reason why the eigenfunctions associated with the second eigenvalue $\lambda_1$ of the Laplace-Beltrami operator captures the geometric structure of environments, as illustrated in the experiments below.

**Theorem 1** *[4]: Define $\lambda_1$ to be the first (non-zero) eigenvalue of the Laplace-Beltrami operator $\mathcal{L}$ on a*

graph $G$. Let $h_G$ denote the Cheeger constant of $G$. Then, we have $2h_G \geq \lambda_1$.

## 5 Hodge Theory

In this section we briefly discuss how the graph Laplacian can be generalized to the setting of Riemannian manifolds [17]. Manifolds are *locally Euclidean* sets, defining a bijection (one-to-one, onto mapping) from an open set containing any element $p \in \mathcal{M}$ to the $n$-dimensional Euclidean space. *Smooth* manifolds in addition require that the homeomorphism mapping any point $p$ to its *coordinates* $(\rho_1(p), \ldots, \rho_n(p))$ be a differentiable function with a differentiable inverse. Given two coordinate functions $\rho(p)$ and $\xi(p)$, which are individually called *charts*, the induced mapping $\psi : \rho \circ \xi^{-1} : \mathcal{R}^n \to \mathcal{R}^n$ must have continuous partial derivatives of all orders. *Riemannian* manifolds are smooth manifolds where the Riemann metric defines the notion of length. Given any element $p \in \mathcal{M}$, the *tangent space* $T_p(\mathcal{M})$ is an $n$-dimensional vector space that is isomorphic to $\mathcal{R}^n$. More formally, a Riemannian manifold is a smooth manifold $\mathcal{M}$ with a family of smoothly varying positive definite inner products $g_p, p \in \mathcal{M}$ where $g_p : T_p(\mathcal{M}) \times T_p(\mathcal{M}) \to \mathcal{R}$. For the Euclidean space $\mathcal{R}^n$, the tangent space $T_p(\mathcal{M})$ is clearly isomorphic to $\mathcal{R}^n$ itself. Consequently, the Riemannian inner product is simply $g(x, y) = \langle x, y \rangle_{\mathcal{R}^n} = \sum_i x_i y_i$, which remains the same over the entire space.

Hodge's theorem states that any smooth function on a compact manifold has a discrete spectrum [17]. We are interested in those functions $f$ that are *eigenfunctions* of $\Delta$, the Laplace-Beltrami operator. On the manifold $\mathcal{R}^n$, the Laplace-Beltrami operator is simply $\mathcal{L} = -\sum_i \frac{\partial^2}{\partial x_i^2}$ (the sign is a convention). Eigenfunctions gives us back the function itself, scaled by an eigenvalue. If the domain is the unit circle, the trigonometric functions $\sin(\theta)$ and $\cos(\theta)$ form eigenfunctions, which leads to *Fourier* analysis, the most well-known example of harmonic analysis. Abstract harmonic analysis generalizes Fourier methods to orthogonal decompositions of smooth functions on arbitrary Riemannian manifolds. The *smoothness functional* for an arbitrary real-valued function on the manifold $f : \mathcal{M} \to \mathcal{R}$ is given by

$$S(f) \equiv \int_\mathcal{M} |\nabla f|^2 \, d\mu = \int_\mathcal{M} f \Delta f d\mu = \langle \Delta f, f \rangle_{\mathcal{L}^2(\mathcal{M})}$$

where $\nabla f$ is the gradient vector field of $f$. For $\mathcal{R}^n$, the gradient vector field is $\nabla f = \sum_i \frac{\partial f}{\partial x_i} \frac{\partial}{\partial x_i}$. For a Riemannian manifold $(\mathcal{M}, g)$, where the Riemannian metric $g$ is used to define distances on manifolds, the Laplace-Beltrami operator is given as

$$\Delta = -\frac{1}{\sqrt{\det\, g}} \sum_{ij} \partial_i \left( \sqrt{\det\, g} \ \ g^{ij} \partial_j \right)$$

where $g$ is the Riemannian metric, $\det g$ is the measure of volume on the manifold, and $\partial_i$ denotes differentiation with respect to the $i^{th}$ coordinate function. This general form is invariant to reparameterization, and reduces to the expression for $\Delta$ given earlier for Euclidean spaces $\mathcal{R}^n$ where $G = I$ and $\det G = 1$.

**Theorem 2** *(Hodge [17]): Let $(\mathcal{M}, g)$ be a compact connected oriented Riemannian manifold. There exists an orthonormal basis for all smooth (square-integrable) functions $\mathcal{L}^2(\mathcal{M}, g)$ consisting of eigenfunctions of the Laplacian. All the eigenvalues are positive, except that zero is an eigenvalue with multiplicity 1.*

Hodge's theorem guarantees that a smooth function $f \in \mathcal{L}^2(M)$ can be expressed as $f(x) = \sum_{i=0}^\infty a_i e_i(x)$ where $e_i$ are the eigenfunctions of $\Delta$, i.e. $\Delta e_i = \lambda_i e_i$. The smoothness $S(e_i) = < \Delta e_i, e_i >_{\mathcal{L}^2(\mathcal{M})} = \lambda_i$. This theorem shows that the eigenfunctions of the Laplace-Beltrami operator can be used to approximate any value function on a manifold. We now turn to describe some experimental results.

## 6 Experiments

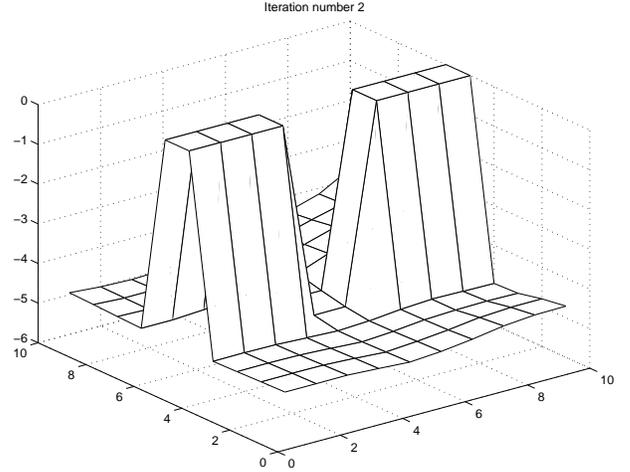

Figure 5: The state value function produced during a run of RPI in a two room gridworld with 100 states.

Figure 5 shows a sample state value function output by RPI for a 100 state grid world MDP partitioned into two room. Here, an initial random walk of 9144 steps was used to learn an undirected graph, from which 20 basis functions were extracted by computing the normalized Laplacian. A constant reward of $-1$ was provided at each step, and the goal is to reach the diagonally opposite corner in the far room (hidden from view). Note how the nonlinear geometry of the environment is transparently reflected in the approximation produced by RPI. Figure 6 gives an example of the eigenfunctions of the Laplacian for a grid world MDP consisting of five rooms, each of dimension $21 \times 20$ connected by a single door in the middle of the boundary walls separating each room. The smoothness of each basis function depends on its spectral ranking: low-order functions are *geodesically* smooth: within each room, they are smooth, but discontinuous across rooms to capture the geometric invariant of the walls separating the rooms. Higher order functions provide harmonic functions for reconstructing value functions within a room. These properties makes Laplacian basis functions well-suited to value function approximation. Figure 7 shows how Laplacian eigenfunctions capture the presence of a centrally placed obstacle in a square grid. Finally, Figure 8 shows how eigenfunctions can efficiently approximate a given value function in a four room grid world (here, the goal is to get to the diagonally opposite corner of the last room).

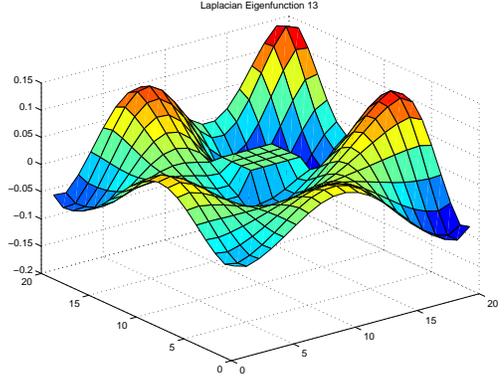

Figure 7: Laplacian eigenfunction for a $20 \times 20$ grid world with a central obstacle. Note how the basis function captures the nonlinearity of the obstacle.

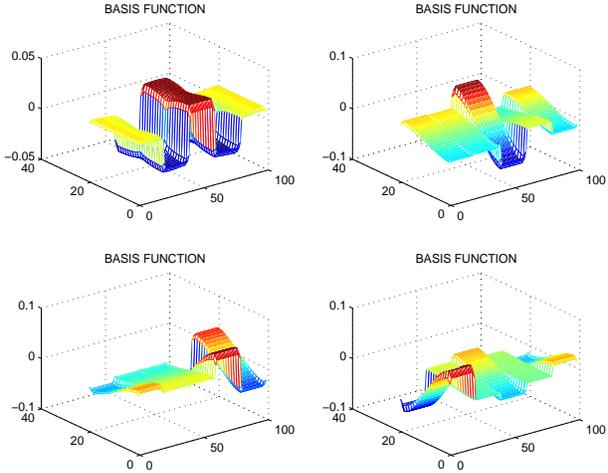

Figure 6: Laplacian eigenfunctions for a five-rooom grid world MDP, where rooms are connected by a single central door. The low-order basis functions capture large-scale geometry.

Next, we provide a detailed control learning experiment comparing the learned basis functions using RPI with using two handcoded basis functions in LSPI, polynomial encoding and radial-basis functions (RBF) (see Table 1). Following [8], +1 rewards are given in states 10 and 41 only in this 50 state chain. Each row reflects the performance of either RPI using learned basis functions or LSPI with a handcoded basis function (values in parentheses indicate the number of basis functions used for each architecture). Each result is the average of five experiments on a sample of $10,000$ transitions. The two numbers reported are average steps to convergence and the average error in the learned policy (number of incorrect actions). The results show the automatically learned Laplacian basis functions in RPI provide a more stable performance at both the low end (5 basis functions) and at the higher end with $k = 25$ basis functions. As the number of basis functions are increased, RPI takes longer to converge, but learns a more accurate policy. LSPI with RBF is unstable at the low end, converging to a very poor policy for 6 basis functions. LSPI with a 5 degree polynomial approximator works reasonably well, but its performance noticeably degrades at higher de-grees, converging to a very poor policy in one step for $k = 15$ and $k = 25$.

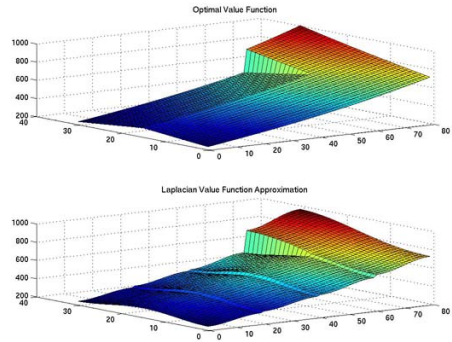

Figure 8: Top: optimal value function. Bottom: approximation using 20 basis functions, both for a four room grid world, where each room is of size $31 \times 20$, achieving a dimensionality reduction from $\mathcal{R}^{2480} \rightarrow \mathcal{R}^{20}$.

| Method | #Trials | Error |
|---|---|---|
| RPI (5) | 4.2 | -3.8 |
| RPI (15) | 7.2 | -3 |
| RPI (25) | 9.4 | -2 |
| LSPI RBF (6) | 3.8 | -20.8 |
| LSPI RBF (14) | 4.4 | -2.8 |
| LSPI RBF (26) | 6.4 | -2.8 |
| LSPI Poly (5) | 4.2 | -4 |
| LSPI Poly (15) | 1 | -34.4 |
| LSPI Poly (25) | 1 | -36 |

Table 1: This table compares the performance of RPI using automatically learned basis functions with LSPI combined with two handcoded basis functions on a 50 state chain graph problem. See text for explanation.

## 7  Discussion

Further development of RPI is continuing on several fronts, including incremental variants, extensions to factored MDPs, and scaling to large state and action spaces. The basis functions used in this paper are global Fourier eigenfunctions. A related approach from harmonic analysis is to use multiscale wavelets, which are based not on differential equations, but dilation equations. *Diffusion wavelets* [5] are a general multiresolution model of diffusion processes on manifolds and graphs. A detailed study of Fourier eigenfunctions and diffusion wavelets with RPI for value function approximation has been completed [12]. The left inverse of the Laplacian is called the *Green's* function and closely relates to the solution of the Bellman equation. We have also developed a novel fast policy evaluation method by using diffusion wavelets to compactly represent powers of the transition matrix [9]. In large graphs the learned basis functions can be computed from samples of the complete graph using *Nystrom* approximations and other low-rank methods.

## Acknowledgments

This research was supported in part by the National Science Foundation under grant ECS-0218125. I thank Mauro Maggioni of the Department of Mathematics at Yale University, and members of the Autonomous Learning Laboratory at the University of Massachusetts, Amherst, for their feedback.